\documentclass[pdflatex,iicol,sn-basic,Numbered]{sn-jnl}

\usepackage{graphicx}
\usepackage{amsmath,amssymb,amsfonts}
\usepackage{array}
\usepackage{textcomp}

\begin{document}

\title[DGV-DLIVO]{DGV-DLIVO: Degeneracy-Guided Visual Fusion for Direct LiDAR-Inertial-Visual Odometry in Degenerate Tunnel Environments}

\author[1]{\fnm{Junrui} \sur{Zhang}}

\affil[1]{\orgdiv{The Department of Electronic and Information Engineering}, \orgname{Tongji University}, \orgaddress{ \city{Shanghai}, \postcode{200092}, \country{China}}}

\abstract{Reliable localization in long underground infrastructures remains a major challenge for autonomous inspection robots. In hydropower spillway tunnels and drainage galleries, repetitive geometric structures provide insufficient constraints along the tunnel axis, while weak texture and illumination variations further degrade visual estimation. Under this coupled degradation, existing LiDAR-inertial and visual-inertial odometry systems may lose the ability to recover continuous longitudinal motion. This paper presents DGV-DLIVO,a degeneracy-guided visual fusion for direct LiDAR-inertial-visual odometry for extreme tunnel environments. Instead of treating vision as a general complementary sensor, the proposed method exploits the LiDAR-degenerate direction as an active guidance signal to determine informative visual regions, establish LiDAR-anchored visual associations, and perform subspace-selective fusion. The visual constraints are introduced only into the deficient motion subspace, while unreliable LiDAR information in the same direction is adaptively suppressed and reliable geometric constraints are preserved. Real-world experiments in hydropower tunnels demonstrate that representative state-of-the-art LIO, LIVO, and VIO methods suffer from stagnation, oscillation, severe under-travel, curved drift, or divergence, whereas DGV-DLIVO maintains continuous and directionally consistent tunnel-axis motion estimation under both cart-mounted and handheld acquisition. It achieves a 91.4\% progress ratio over the 40~m tunnel sequence and significantly improves longitudinal recovery and trajectory stability. These results demonstrate the effectiveness of degeneracy-guided adaptive fusion for robotic inspection in severely degraded underground environments.The source code and a supplementary video are publicly available at \url{https://github.com/0RayZhang0/DGV-DLIVO}.}


\maketitle

\section{Introduction}
\label{sec:introduction}

\begin{figure*}[t]
	\centering
	\includegraphics[width=\textwidth]{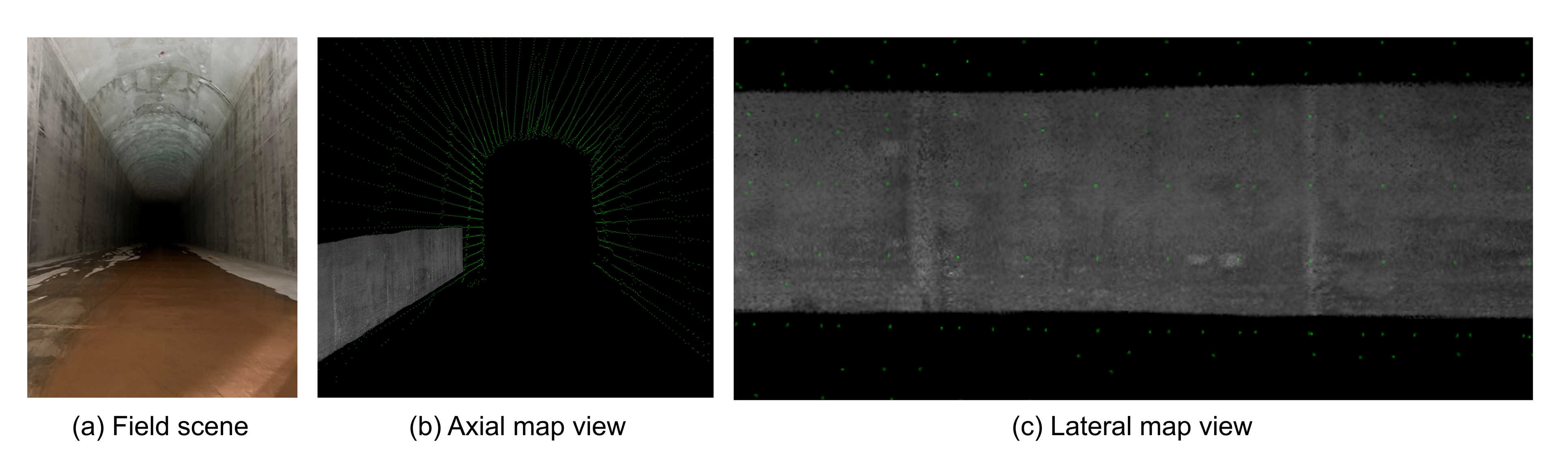}
	\caption{
		Real hydropower spillway-tunnel environment and the reconstruction result of the proposed method.
		(a) Field scene.
		(b) Axial view of the camera-colorized point-cloud map.
		(c) Lateral view of the reconstructed map.
		The proposed method reconstructs the tunnel structure from multimodal measurements, where camera-observed regions are colorized by image information and the remaining points are displayed as point clouds.
	}
	\label{fig:tunnel_environment_map}
\end{figure*}

Large hydropower facilities require routine inspection and operational maintenance of spillway tunnels, drainage galleries, and other internal passages. In large installations, these passages may extend over long distances and often have nearly straight alignments and repeated cross-sectional structures. Mobile inspection robots can support repeatable data collection in these spaces, but continuous and stable pose estimation is a prerequisite for navigation and inspection task execution. Because GNSS is unavailable in such enclosed passages, this localization capability must be provided primarily by onboard odometry. LiDAR-inertial odometry (LIO) is widely used because LiDAR provides illumination-invariant geometric measurements and IMU integration improves short-term motion prediction~\cite{xu2022fastLIO2,chen2023DLIO,shan2020liosam,bai2022fasterlio,he2023pointlio,lee2024lidarOdometrySurvey}. However, its geometric constraints can become insufficient in long, self-similar passages.

The target degeneracy is illustrated in Fig.~\ref{fig:tunnel_environment_map}. The field scene and the reconstructed camera-colorized point-cloud map reveal a long, regular, and nearly invariant tunnel structure. A recent survey identifies feature-scarce and repetitive environments, including tunnels and long corridors, as a major unresolved condition for LiDAR odometry because the absence of distinctive geometry introduces ambiguity into scan matching~\cite{lee2024lidarOdometrySurvey}. In the hydropower passages considered here, the surrounding surfaces strongly constrain motion perpendicular to the tunnel axis but provide little information along the axis. Different longitudinal translations may therefore produce similar registration costs, leading to an ill-conditioned estimation problem, stagnation, or accumulated drift~\cite{guo2026large}. This is a structural degeneracy of the environment and cannot be fully resolved by denser point clouds or more accurate implementation of the same geometric registration objective.

Cameras can provide complementary cues from wall markings, joints, stains, and illumination changes. However, hydropower galleries frequently contain weak texture, repeated appearance, and uneven or insufficient illumination, which also limit standalone visual and visual-inertial odometry~\cite{qin2018vins,campos2021orbslam3,engel2018dso}. A related localization difficulty has been reported for mobile robots operating in repetitive, low-texture warehouse aisles, where it is described as the infinite-corridor problem~\cite{prasad2025wareOdo}. WareOdo mitigates this problem using lane structures and artificial fiducial markers. Such landmarks can provide useful global corrections, but they do not eliminate the need for continuous odometry between landmark observations. In the present work, the sparse wall markers used during the experiments serve only as evaluation references and are not measurements available to the estimator.

General LiDAR-visual and LiDAR-visual-inertial frameworks improve robustness by combining complementary sensing modalities~\cite{shan2021lvisam,lin2021r2live,lin2022r3live,zheng2024fastLIVO2}. For example, Meng~et~al.~\cite{meng2022tightlyCoupledVLO} jointly optimize depth-recovered visual features and LiDAR features and further employ loop closure for globally consistent estimation. These approaches demonstrate the benefit of visual--LiDAR fusion, but their measurements are generally applied as full-state constraints or depend on a visual subsystem that must remain independently reliable. Under simultaneous LiDAR degeneracy and repetitive visual appearance, generic fusion may fail to recover the missing tunnel-direction motion and may allow uncertain visual information to disturb directions that remain well constrained by LiDAR.

To address this challenge, this work explores a different perspective from conventional multimodal odometry. Instead of adding a generic visual module to an existing LIO or LIVO pipeline, the proposed method uses the LiDAR-degenerate direction to guide visual information selection, visual association, and subspace fusion. Visual cues are used to compensate the LiDAR-deficient motion component, while LiDAR remains dominant in well-observed directions. The objective is to provide a stable odometry basis for mobile inspection robots in long hydropower passages

The contributions of this work are as follows:
\begin{enumerate}
	\item We introduce DGV-DLIVO, a degeneracy-guided visual fusion for direct LiDAR-inertial-visual odometry framework specifically designed for long hydropower passages with severe tunnel-axis geometric degeneracy. Unlike conventional multimodal odometry systems that use visual measurements as general complementary information, DGV-DLIVO exploits the detected LiDAR weak direction as a unified guidance signal for visual constraint construction and multimodal state estimation.
	
	\item We develop a degeneracy-guided visual constraint construction strategy that selects visual information according to its contribution to recovering the LiDAR-deficient motion direction. By coupling selected image patches with locally verified LiDAR geometry, the method establishes metric visual constraints with persistent association capability under sparse texture, repetitive structures, and illumination variations.
	
	\item We propose a degeneracy-adaptive subspace-selective fusion strategy that dynamically reallocates the contributions of LiDAR and visual measurements according to geometric observability. Visual residuals are introduced only into the deficient motion subspace, while unreliable LiDAR components in the same direction are suppressed and reliable geometric constraints are preserved in the complementary subspace.
	
	\item We validate DGV-DLIVO through extensive real-world experiments in cart-mounted and handheld hydropower tunnels. Compared with representative state-of-the-art LIO, LIVO, and VIO methods, DGV-DLIVO maintains continuous and accurate tunnel-axis motion estimation under severe geometric and visual degradation, whereas existing approaches exhibit stagnation, oscillation, severe under-travel, drift, or divergence. 
\end{enumerate}

To benefit the robotics community, we publicly release the source
code of DGV-DLIVO to support further research on robust localization
in degenerate environments.

\begin{figure*}[t]
	\centering
	\includegraphics[width=\textwidth]{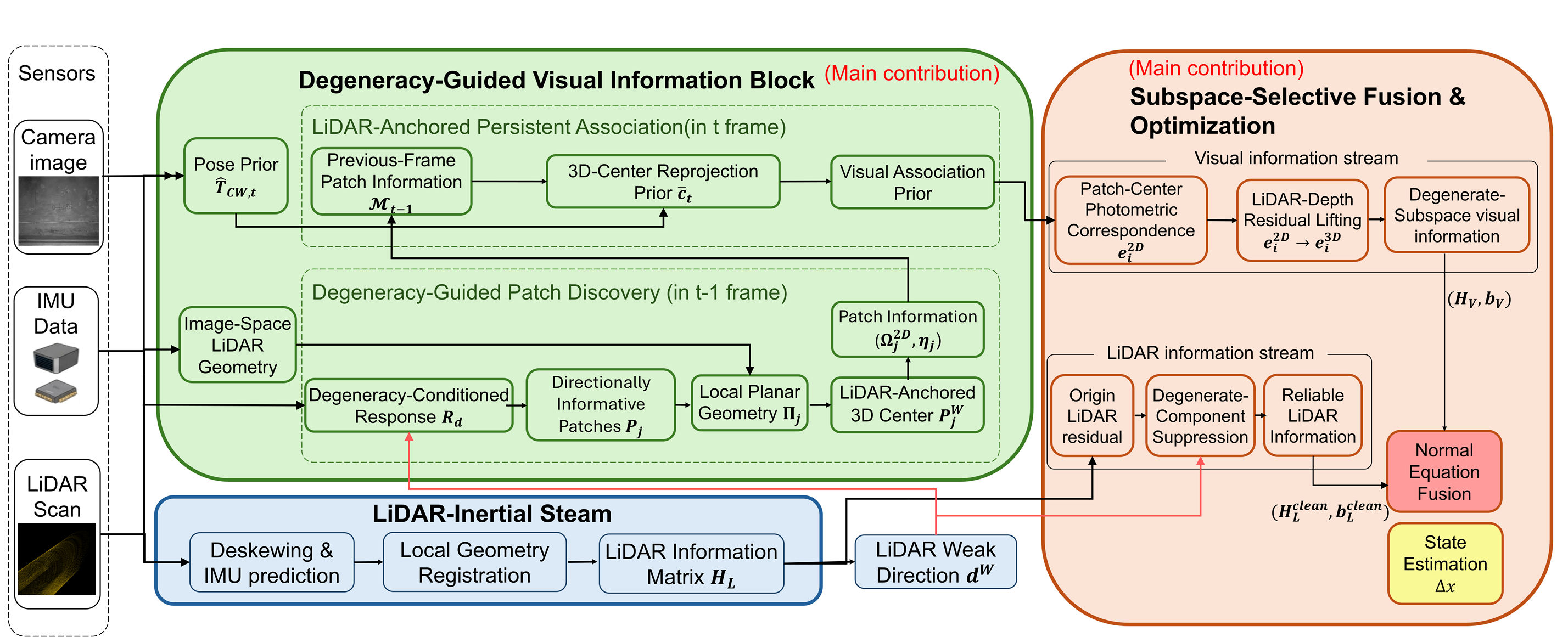}
	\caption{
		System overview
	}
	\label{fig:system_overview}
\end{figure*}

\section{Related Work}
\label{sec:related_work}

\subsection{LiDAR-Inertial Odometry}
\label{sec:rw_lio}

LiDAR odometry has evolved from the feature-based two-thread pipeline of LOAM~\cite{zhang2014loam} to direct registration methods~\cite{vizzo2023kissicp} and tightly coupled LiDAR-inertial systems. FAST-LIO2~\cite{xu2022fastLIO2}, DLIO~\cite{chen2023DLIO}, Faster-LIO~\cite{bai2022fasterlio}, and Point-LIO~\cite{he2023pointlio} improve filtering, registration, map maintenance, and update efficiency in different ways. These methods provide strong geometric odometry backbones, but their constraints remain purely geometric and can become rank-deficient in self-similar structures.

\subsection{LiDAR-Visual-Inertial Fusion}
\label{sec:rw_lvi}

To compensate for the limitations of geometry-only sensing, a variety of systems fuse cameras with LiDAR and IMU. LIC-Fusion~\cite{zuo2019licfusion} and its extension~\cite{zuo2020licfusion2} integrate camera features, LiDAR plane constraints, and IMU in a multi-state Kalman filter, while LVI-SAM~\cite{shan2021lvisam} adds visual-inertial factors to the factor-graph structure of LIO-SAM~\cite{shan2020liosam}. R2LIVE~\cite{lin2021r2live} and R3LIVE~\cite{lin2022r3live} unify all three modalities in a tightly coupled estimator and produce textured point-cloud maps.  These systems demonstrate the general benefit of multi-modal fusion but assume that both the visual and LiDAR subsystems can independently produce usable state updates.

The FAST-LIVO line of work~\cite{zheng2022fastLIVO1, zheng2024fastLIVO2} achieves high efficiency by combining sparse-direct photometric residuals with LiDAR geometric residuals in an iterated Kalman filter. Because FAST-LIVO2 anchors visual patches to LiDAR map points, its visual observations are spatially tied to the LiDAR geometry. In weakly textured tunnel scenes, LiDAR points may project onto featureless image regions, causing the system to miss the few informative visual cues available elsewhere in the image. LVIO-Fusion~\cite{zhang2024lviofusion} targets LiDAR-degenerate and texture-less scenes by initializing VIO from LIO output and refining the state with coarse-to-fine photometric errors, but its visual refinement acts on the full state rather than being conditioned on the LiDAR-degenerate direction.

Domain-specific systems such as MetroLoc~\cite{wang2025metroloc} and tunnel LiDAR-inertial mapping~\cite{dai2023tunnelLidarInertial} further show the severity of geometric degeneracy in underground structures, but they rely on domain priors that are not directly transferable to general tunnel or corridor geometries.

\subsection{Degeneracy-Aware Methods and Complementary Fusion}
\label{sec:rw_degen}

A growing line of research recognizes that LiDAR registration degeneracy is directional and that compensation should target the deficient subspace rather than the full state. X-ICP~\cite{tuna2024xicp} analyzes per-direction localizability in ICP and suppresses updates along weakly constrained directions. DALI-SLAM~\cite{wu2025dali} incorporates degeneracy-aware processing into a LiDAR-inertial SLAM pipeline with multi-constraint pose-graph optimization.  Both operate within the LiDAR-inertial domain without introducing an external sensing modality.

Repetitive and weakly textured passages also challenge inspection-robot localization. WareOdo~\cite{prasad2025wareOdo} uses lane markings and fiducial markers to correct odometry in repetitive warehouse aisles. Such artificial landmarks are generally unsuitable for hydropower spillway tunnels, which must remain unobstructed to convey large water flows during operation. Localization must therefore rely on features already present in the environment. Our method detects these natural image structures and anchors them metrically using local LiDAR geometry.

Several methods inject complementary observations into the degenerate subspace. COIN-LIO~\cite{pfreundschuh2024coin} constructs LiDAR intensity images and selects features complementary to geometric constraints, but its performance depends on reflectance quality and scan density. DR-LRIO~\cite{nissov2024drlrio} uses radar Doppler velocity for complementary motion information, although its loose-coupling formulation does not explicitly remove unreliable LiDAR components from the degenerate subspace. LOFF~\cite{zhang2024loff} compensates wall-induced degeneracy with optical-flow odometry, but its detection relies on geometric heuristics rather than Hessian eigendecomposition.

Switch-SLAM~\cite{lee2024switch} runs LIO and visual SLAM in parallel and switches between them according to degeneracy indicators, requiring each subsystem to remain usable at least intermittently. Yan~et~al.~\cite{yan2026deepTunnelLVI} selectively fuse VIO under coupled rotation-translation degeneracy with deep visual features, but require a full VIO front-end and GPU-equipped hardware. In contrast, our method provides lightweight image-level compensation only for the LiDAR-deficient subspace.


\section{Method}
\label{sec:method}
\subsection{System Overview}
\label{subsec:system_overview}

Fig.~\ref{fig:system_overview} summarizes the proposed degeneracy-guided visual compensation framework. The LiDAR-inertial stream, instantiated using DLIO~\cite{chen2023DLIO}, provides an inertial pose prior, the LiDAR normal equations $(H_L,b_L)$, and a weak direction $d^W$ identified from the eigendecomposition of the LiDAR information matrix~\cite{tuna2024xicp}. Rather than serving only as a degeneracy indicator, $d^W$ jointly conditions visual information discovery and the subsequent normal-equation update.

The visual constraint is constructed through two coupled processes. First, the weak direction conditions a full-image directional response for selecting patches informative to the LiDAR-deficient motion component. Image-space LiDAR geometry is then used to verify local planar consistency and establish a metric 3D center for each retained patch. Second, the IMU-predicted pose reprojects the previous-frame 3D center to provide an association prior, which is refined through patch-center photometric correspondence. The resulting 2D residual is lifted using the LiDAR-derived metric depth and restricted to the deficient subspace. Meanwhile, unreliable components of the LiDAR normal equations in that subspace are suppressed before the complementary visual and LiDAR information is fused.

Therefore, the camera is not used as an independent full-state odometry source. Instead, it provides a directionally targeted constraint for the motion component that is structurally missing from LiDAR registration, while LiDAR remains dominant in well-conditioned directions.

\begin{figure}[t]
	\centering
	\includegraphics[width=\linewidth]{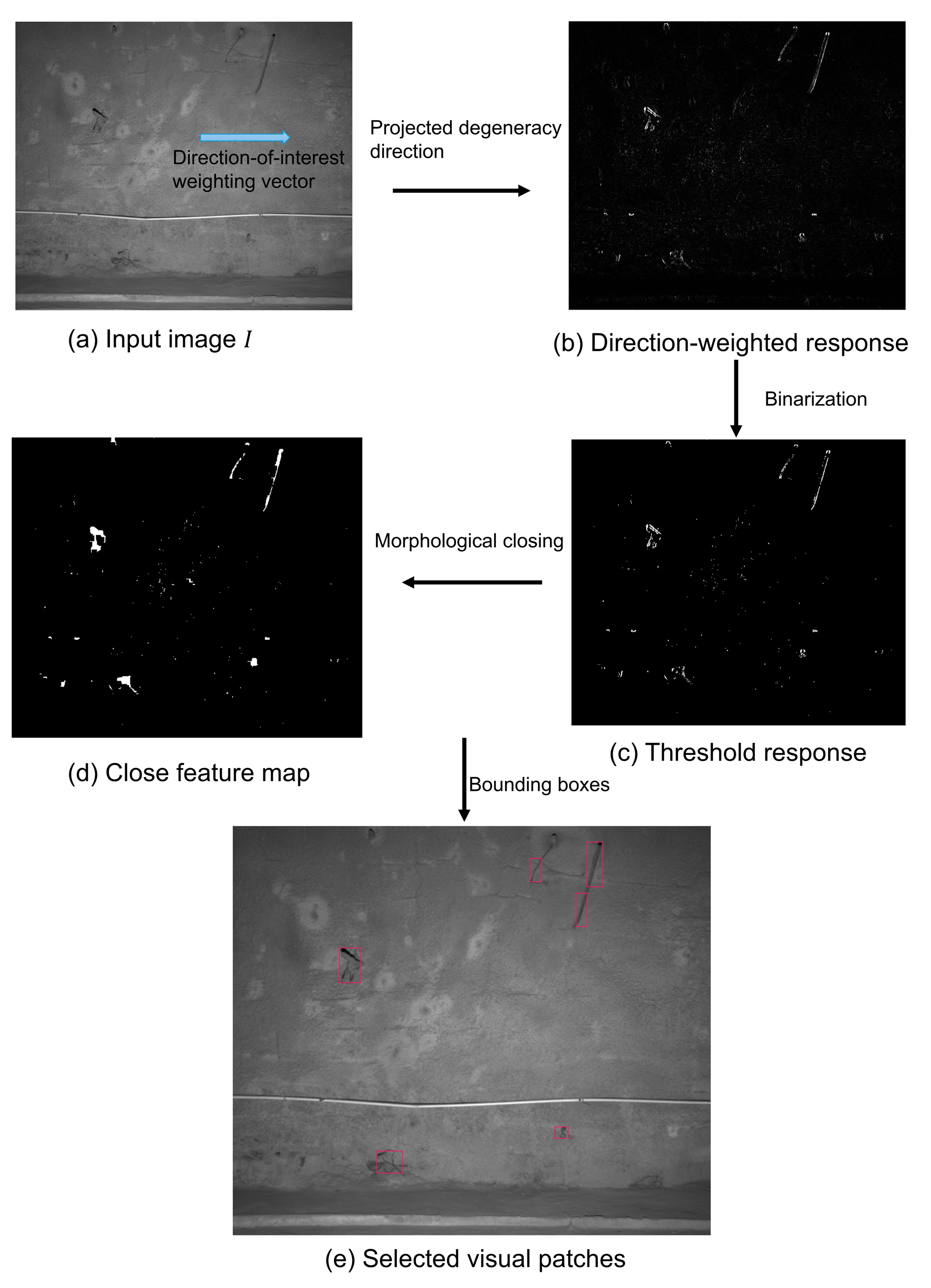}
	\caption{
		Degeneracy-guided visual patch selection.
		(a) The input image and the projected degeneracy direction define the image-plane direction of interest.
		(b) A direction-weighted image response emphasizes intensity variations along this direction.
		(c) Thresholding suppresses weak directional responses.
		(d) Morphological closing produces spatially coherent support regions.
		(e) Bounding boxes on the original input image define the selected visual patches used for tracking and optimization
	}
	\label{fig:degeneracy_guided_visual_patch_selection}
\end{figure}

\subsection{Degeneracy-Guided Visual Patch Information}
\label{subsec:degeneracy_guided_visual_patch_information}

This module extracts visual patches informative for the LiDAR-degenerate subspace. In weak visual-feature scenes, useful structures are often sparse and unevenly distributed. Searching only near projected LiDAR points, as in local-map-guided direct LIVO association~\cite{zheng2024fastLIVO2}, may miss usable gradients elsewhere in the image. Inspired by FAST-LIVO2~\cite{zheng2024fastLIVO2} and COIN-LIO~\cite{pfreundschuh2024coin}, we perform full-image patch selection and adapt the image direction of interest according to the detected LiDAR degeneracy. Each patch aggregates local gradients over an adaptive region, making the measurement sensitive to both point-like and line-like structures.

Let $I$ denote the original grayscale input image shown in Fig.~\ref{fig:degeneracy_guided_visual_patch_selection}(a), and let $d^W$ be the detected degenerate direction in the world frame. The direction is first transformed into the camera frame by
\begin{equation}
	d^C = R_{CW} d^W .
	\label{eq:patch_deg_dir_camera}
\end{equation}
Its image-plane direction of interest is defined as
\begin{equation}
	w =
	\frac{
		\begin{bmatrix}
			f_x d_x^C\\
			f_y d_y^C
	\end{bmatrix}}
	{\left\|
		\begin{bmatrix}
			f_x d_x^C\\
			f_y d_y^C
		\end{bmatrix}
		\right\|},
	\label{eq:patch_interest_direction}
\end{equation}
where $f_x$ and $f_y$ are the camera focal lengths, and $w=[w_u,w_v]^\top$ represents the projected compensation direction on the image plane.

To construct the direction-sensitive response in Fig.~\ref{fig:degeneracy_guided_visual_patch_selection}(b), the horizontal and vertical Scharr derivatives of the original image are computed as
\begin{equation}
	I_u = K_u * I, \qquad
	I_v = K_v * I ,
	\label{eq:patch_scharr_gradients}
\end{equation}
where $K_u$ and $K_v$ are the horizontal and vertical Scharr kernels. The degeneracy-induced directional image kernel is
\begin{equation}
	K_d(w)=w_uK_u+w_vK_v .
	\label{eq:patch_directional_kernel}
\end{equation}
The corresponding directional response map is then
\begin{equation}
	R_d(x)=\left|(K_d(w)*I)(x)\right|
	=\left|w_uI_u(x)+w_vI_v(x)\right| .
	\label{eq:patch_directional_response}
\end{equation}
This response emphasizes image structures whose intensity variation is informative along the projected LiDAR-degenerate direction, rather than selecting generic corners independent of the estimator's deficient subspace.

The response map is converted into spatial support regions by thresholding the normalized response, as shown in Fig.~\ref{fig:degeneracy_guided_visual_patch_selection}(c), followed by morphological closing, as shown in Fig.~\ref{fig:degeneracy_guided_visual_patch_selection}(d):
\begin{equation}
	M_d =
	\mathrm{Close}
	\left(
	\mathbf{1}[\mathcal{N}(R_d)>\tau_b];
	\mathcal{S}
	\right),
	\label{eq:patch_support_mask}
\end{equation}
where $\mathcal{N}(\cdot)$ denotes normalization, $\tau_b$ is the response threshold, and $\mathrm{Close}(\cdot;\mathcal{S})$ is the closing operation with structuring element $\mathcal{S}$. Candidate patches are obtained from connected support regions:
\begin{equation}
	\mathcal{C} = \mathrm{CC}(M_d),
	\qquad
	P_j = \mathrm{bbox}(C_j),\; C_j \in \mathcal{C}.
	\label{eq:patch_connected_components}
\end{equation}
The response map only localizes informative regions, while each final patch $P_j$ and its visual information are evaluated on the original image, as illustrated in Fig.~\ref{fig:degeneracy_guided_visual_patch_selection}(e).

For each retained patch, let $I_j=I|_{P_j}$ denote the corresponding image crop. Its local image gradients are
\begin{equation}
	I_{j,u}=K_u*I_j,\qquad I_{j,v}=K_v*I_j .
	\label{eq:patch_local_gradients}
\end{equation}
For a pixel $\xi$ inside $P_j$, the gradient vector is defined as
\begin{equation}
	g_j(\xi)=
	\begin{bmatrix}
		I_{j,u}(\xi)\\
		I_{j,v}(\xi)
	\end{bmatrix}.
	\label{eq:patch_gradient_vector}
\end{equation}
The two-dimensional structural information matrix of the patch is then computed by
\begin{equation}
	\Omega_j^{2D}
	=
	\frac{1}{|P_j|s_g^2}
	\sum_{\xi\in P_j}
	g_j(\xi)g_j(\xi)^\top
	+
	\epsilon I_2 ,
	\label{eq:patch_2d_information}
\end{equation}
where $s_g$ normalizes the gradient scale, and $\epsilon I_2$ ensures positive definiteness.

The directional observability of patch $P_j$ is evaluated by projecting this structural information matrix onto the image-plane direction of interest:
\begin{equation}
	\eta_j = w^\top \Omega_j^{2D} w .
	\label{eq:patch_directional_score}
\end{equation}
Patches with insufficient $\eta_j$ are excluded, since they provide weak visual information along the LiDAR-degenerate direction. The retained $\Omega_j^{2D}$ is later lifted through the projection Jacobian to form the depth-anchored three-dimensional visual information matrix used in subspace-selective fusion.

\begin{figure}[t]
	\centering
	\includegraphics[width=\linewidth]{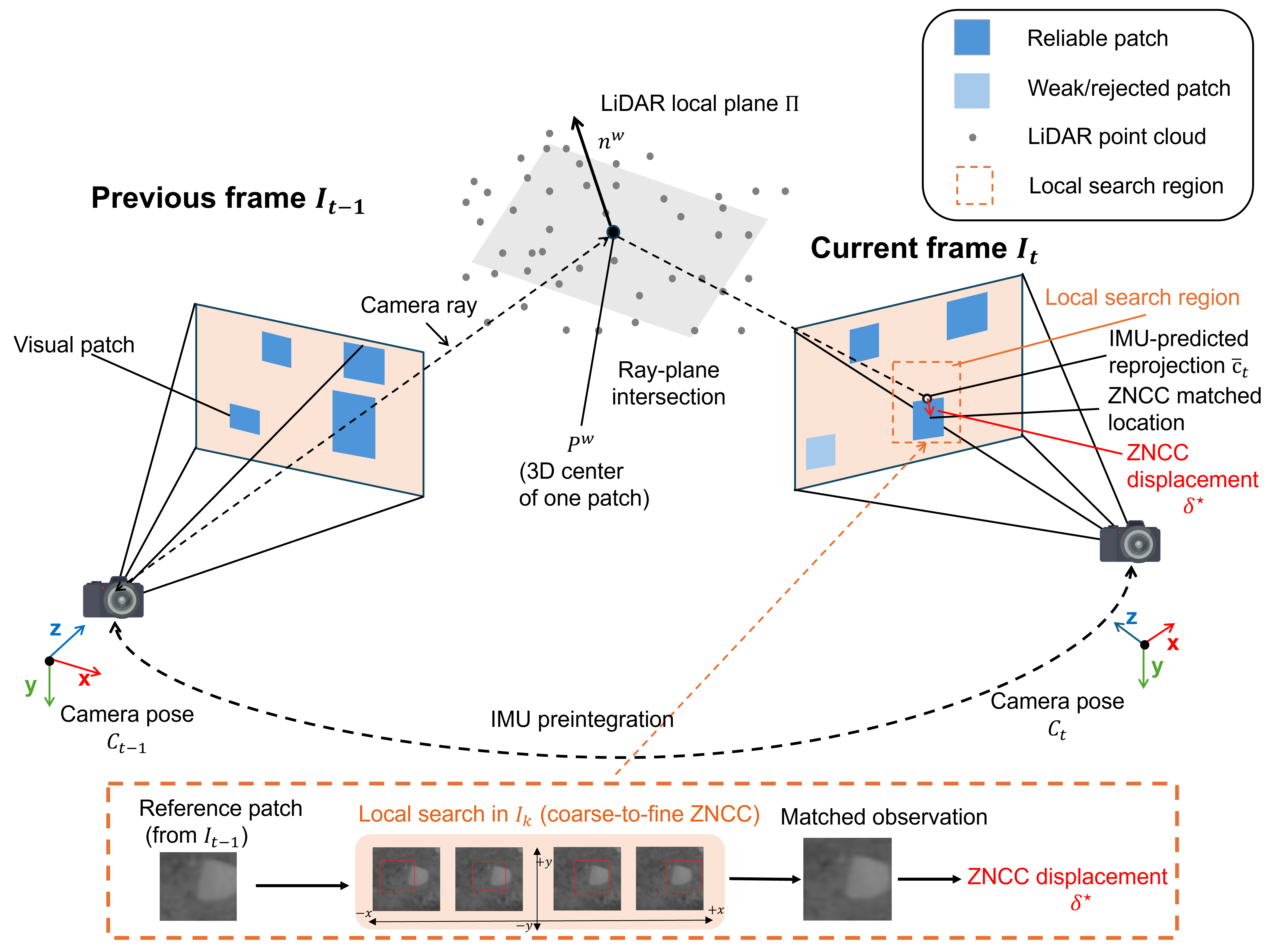}
	\caption{
		LiDAR-anchored persistent visual patch tracking.
		A selected patch in $I_{t-1}$ is assigned a 3D center by intersecting its camera ray with a local LiDAR plane.
		The IMU-predicted camera pose reprojects this center to $\bar c_t$ in $I_t$, and a local coarse-to-fine ZNCC search finds the matched location $z_t$.
		The matched location is used to define a pose-dependent reprojection residual for subspace-selective fusion
	}
	\label{fig:lidar_anchored_patch_tracking}
\end{figure}

\subsection{LiDAR-Anchored Persistent Patch Tracking}
\label{subsec:lidar_anchored_tracking}

This module maintains visual patch associations in weak-texture and repetitive scenes. Instead of tracking isolated points or matching descriptors globally, each patch is anchored by local LiDAR geometry, as shown in Fig.~\ref{fig:lidar_anchored_patch_tracking}. For one patch, the cached state from the previous frame is
\begin{equation}
	\mathcal{M}_{t-1}
	=
	\left\{
	P^W,\,
	B_{t-1},\,
	I_{t-1}
	\right\},
	\label{eq:tracking_patch_state}
\end{equation}
where $P^W$ is the LiDAR-anchored 3D center, $B_{t-1}$ is the image support region, and $I_{t-1}$ is the reference patch appearance. 

The metric depth of each patch center is first provided by the local LiDAR surface. Let $\Pi$ denote the local plane fitted from LiDAR points projected near the patch region, with normal $n^W$ and a point $\bar p^W$ on the plane. For the patch center pixel in $I_{t-1}$, let $O^W$ be the camera center and $r^W$ be the corresponding camera ray in the world frame. The LiDAR-anchored 3D center is obtained by ray-plane intersection:
\begin{equation}
	\lambda
	=
	\frac{(n^W)^\top(\bar p^W-O^W)}
	{(n^W)^\top r^W},
	\qquad
	P^W
	=
	O^W+\lambda r^W .
	\label{eq:tracking_ray_plane_center}
\end{equation}
Thus, each visual patch is associated with a metric 3D anchor supplied by the local LiDAR geometry.

Given the IMU-predicted camera pose $\hat T_{CW,t}$, the 3D patch center is reprojected into the current image as
\begin{equation}
	\bar c_t
	=
	\pi\!\left(\hat T_{CW,t}P^W\right).
	\label{eq:tracking_predicted_center}
\end{equation}
The predicted center $\bar c_t$ defines a local search region rather than a final visual observation. This constrains patch association by the inertial motion prior and reduces ambiguity in weak-texture or repetitive image regions.

Around $\bar c_t$, the current patch correspondence is estimated by coarse-to-fine ZNCC search. Let $\mathcal{S}$ be the local search region and $\rho_{\mathrm{ZNCC}}(\cdot)$ denote the zero-mean normalized cross-correlation score between the reference patch in $I_{t-1}$ and a candidate patch in $I_t$. The best displacement is
\begin{equation}
	\delta^\star
	=
	\arg\max_{\delta\in\mathcal{S}}
	\rho_{\mathrm{ZNCC}}
	\!\left(
	I_{t-1},
	I_t(\bar c_t+\delta)
	\right).
	\label{eq:tracking_zncc_displacement}
\end{equation}
The matched image location is then
\begin{equation}
	z_t
	=
	\bar c_t+\delta^\star .
	\label{eq:tracking_matched_location}
\end{equation}
Here, the ZNCC result provides the image correspondence of the LiDAR-anchored patch center, rather than an independent pose correction.

Let $\Delta T$ be the pose correction left-multiplied to the IMU-predicted camera pose. The corrected camera pose is
\begin{equation}
	T_{CW,t}(\Delta T)
	=
	\Delta T \hat T_{CW,t}.
	\label{eq:tracking_corrected_pose}
\end{equation}
The resulting visual reprojection residual is defined as
\begin{equation}
	e_t^{2D}(\Delta T)
	=
	\pi\!\left(T_{CW,t}(\Delta T)P^W\right)
	-
	z_t .
	\label{eq:tracking_visual_residual}
\end{equation}
When the pose correction is accurate, the projection of the LiDAR-anchored 3D center coincides with the ZNCC correspondence. This residual, together with the 3D anchor $P^W$ and the patch information matrix from Sec.~\ref{subsec:degeneracy_guided_visual_patch_information}, is passed to Sec.~\ref{subsec:subspace_fusion} for depth-anchored visual constraint construction.

\subsection{Subspace-Selective Normal Equation Fusion}
\label{subsec:subspace_fusion}

The role of this module is to compensate the LiDAR-degenerate subspace at the normal-equation level. For each valid visual patch, Sec.~\ref{subsec:lidar_anchored_tracking} provides a pose-dependent 2D reprojection residual $e_i^{2D}=[e_{u,i},e_{v,i}]^\top$ and a LiDAR-anchored 3D center $P_i^W$. At the current linearization point, the anchor point in the camera frame is
\begin{equation}
	P_i^C
	=
	T_{CW,t}P_i^W
	=
	[X_i,Y_i,Z_i]^\top .
	\label{eq:fusion_anchor_camera_point}
\end{equation}
Using the LiDAR-anchored depth $Z_i$, the image residual is lifted to a pseudo-3D residual in the camera frame:
\begin{equation}
	e_i^{3D,C}
	=
	\begin{bmatrix}
		Z_i e_{u,i}/f_x\\
		Z_i e_{v,i}/f_y\\
		0
	\end{bmatrix},
	\qquad
	e_i^{3D,W}=R_{WC,t}e_i^{3D,C}.
	\label{eq:fusion_depth_anchored_residual}
\end{equation}
This lifting uses metric depth from the local LiDAR surface rather than monocular triangulation.

The lifted residual is not treated as an unconstrained 3D visual measurement. To suppress errors from patch-center approximation and local matching, we follow the covariance-weighted view of GICP~\cite{segal2009GICP} and restrict the visual information according to the LiDAR-degenerate geometry. The projection Jacobian at the LiDAR-anchored point is
\begin{equation}
	J_\pi(P_i^C)
	=
	\begin{bmatrix}
		f_x/Z_i & 0 & -f_xX_i/Z_i^2\\
		0 & f_y/Z_i & -f_yY_i/Z_i^2
	\end{bmatrix}.
	\label{eq:fusion_projection_jacobian}
\end{equation}
In the one-direction degeneracy mode, the visual information strength along the detected weak direction $d^W$ is evaluated by
\begin{equation}
	\alpha_i
	=
	\left(J_\pi(P_i^C)R_{CW,t}d^W\right)^\top
	\Omega_i^{2D}
	\left(J_\pi(P_i^C)R_{CW,t}d^W\right),
	\label{eq:fusion_directional_visual_strength}
\end{equation}
where $\Omega_i^{2D}$ is the patch structural information matrix from Sec.~\ref{subsec:degeneracy_guided_visual_patch_information}. The corresponding 3D visual information matrix is
\begin{equation}
	\Omega_i^{3D}
	=
	\alpha_i d^W(d^W)^\top .
	\label{eq:fusion_3d_visual_information}
\end{equation}
This rank-one matrix makes the visual term contribute only along the LiDAR-degenerate direction.

In the full-attention mode, the visual information is allowed to act on the local LiDAR tangent plane. Let $n_i^W$ be the local plane normal and define
\begin{equation}
	\Pi_i
	=
	I_3-n_i^W(n_i^W)^\top ,
	\qquad
	A_i
	=
	J_\pi(P_i^C)R_{CW,t}.
	\label{eq:fusion_tangent_lift_operator}
\end{equation}
The lifted visual information matrix is then
\begin{equation}
	\Omega_i^{3D}
	=
	\Pi_i A_i^\top \Omega_i^{2D} A_i \Pi_i
	+
	\epsilon_n n_i^W(n_i^W)^\top .
	\label{eq:fusion_full_attention_information}
\end{equation}
Here, $\epsilon_n$ is a small regularization term used to avoid singular information matrices.

The visual residual contributes to the pose update through the 3D residual Jacobian
\begin{equation}
	J_i^V
	=
	\begin{bmatrix}
		[T_{CW,t}P_i^W]_\times & -I_3
	\end{bmatrix}.
	\label{eq:fusion_visual_jacobian}
\end{equation}
Only the state columns associated with the deficient motion component are retained. For translation degeneracy,
\begin{equation}
	\tilde J_i^V
	=
	\begin{bmatrix}
		0_{3\times3} & -I_3
	\end{bmatrix},
	\label{eq:fusion_visual_jacobian_translation}
\end{equation}
while for rotation degeneracy,
\begin{equation}
	\tilde J_i^V
	=
	\begin{bmatrix}
		[T_{CW,t}P_i^W]_\times & 0_{3\times3}
	\end{bmatrix}.
	\label{eq:fusion_visual_jacobian_rotation}
\end{equation}
The visual normal equations are
\begin{equation}
	H_V
	=
	\sum_i
	(\tilde J_i^V)^\top
	\Omega_i^{3D}
	\tilde J_i^V,
	\qquad
	b_V
	=
	\sum_i
	(\tilde J_i^V)^\top
	\Omega_i^{3D}
	e_i^{3D,W}.
	\label{eq:fusion_visual_normal_equation}
\end{equation}

Due to LiDAR measurement noise and residual geometric associations, the LiDAR residuals and gradients along the degenerate direction are generally not exactly zero even in structurally degenerate scenes. Directly fusing these unreliable components with the visual terms can disturb pose estimation. Therefore, before fusion, the LiDAR normal equations are explicitly projected onto the reliable subspace. Let $d^W$ denote the detected LiDAR-degenerate direction. The degenerate and reliable projectors are
\begin{equation}
	\mathcal Q_d
	=
	d^W(d^W)^\top,
	\qquad
	\mathcal Q_L
	=
	I_3-\mathcal Q_d .
	\label{eq:fusion_subspace_projectors}
\end{equation}
For translation degeneracy, the LiDAR selector is
\begin{equation}
	\mathcal S_t
	=
	\begin{bmatrix}
		I_3 & 0\\
		0 & \mathcal Q_L
	\end{bmatrix},
	\qquad
	\bar H_L=\mathcal S_t^\top H_L\mathcal S_t,
	\qquad
	\bar b_L=\mathcal S_t^\top b_L .
	\label{eq:fusion_lidar_translation_projection}
\end{equation}
For rotation degeneracy, the selector becomes
\begin{equation}
	\mathcal S_r
	=
	\begin{bmatrix}
		\mathcal Q_L & 0\\
		0 & I_3
	\end{bmatrix},
	\qquad
	\bar H_L=\mathcal S_r^\top H_L\mathcal S_r,
	\qquad
	\bar b_L=\mathcal S_r^\top b_L .
	\label{eq:fusion_lidar_rotation_projection}
\end{equation}
This operation removes the unreliable LiDAR Hessian and gradient components in the degenerate subspace, while preserving LiDAR constraints in well-conditioned directions.

The visual weight aligns the scale of the visual normal equations with the reliable LiDAR normal equations:
\begin{equation}
	\gamma
	=
	\min
	\left(
	\frac{\lambda_{\mathrm{nd}}}
	{\mathrm{tr}(H_V)},
	\gamma_{\max}
	\right),
	\label{eq:fusion_visual_weight}
\end{equation}
where $\lambda_{\mathrm{nd}}$ is chosen from the eigenvalue magnitude of the non-degenerate LiDAR subspace, and $\gamma_{\max}$ limits excessive visual influence. In practice, a constant multiple of this LiDAR information scale is sufficient, and the estimator is not sensitive to its exact value.

The fused normal equations are
\begin{equation}
	H
	=
	\bar H_L+\gamma H_V,
	\qquad
	b
	=
	\bar b_L+\gamma b_V .
	\label{eq:fusion_final_normal_equation}
\end{equation}
The pose increment is solved with damping as
\begin{equation}
	(H+\mu I)\Delta x=-b,
	\qquad
	T\leftarrow\exp(\Delta x)T .
	\label{eq:fusion_pose_update}
\end{equation}
The resulting estimator remains LiDAR-dominant in well-conditioned directions and uses visual information only in the identified degenerate subspace.

\section{Experiments}
\label{sec:experiments}

The target application is routine robotic inspection and operational maintenance in large hydropower facilities. Spillway tunnels and drainage galleries can form long, nearly straight passages with highly repetitive cross-sections. Continuous localization is therefore a prerequisite for reliable robot navigation and inspection task execution in these spaces. The experiments reported here evaluate the odometry and localization component under this target degeneracy.

\subsection{Experimental Setup}
\label{subsec:experimental_setup}

\begin{figure}[t]
	\centering
	\includegraphics[width=0.8\columnwidth]{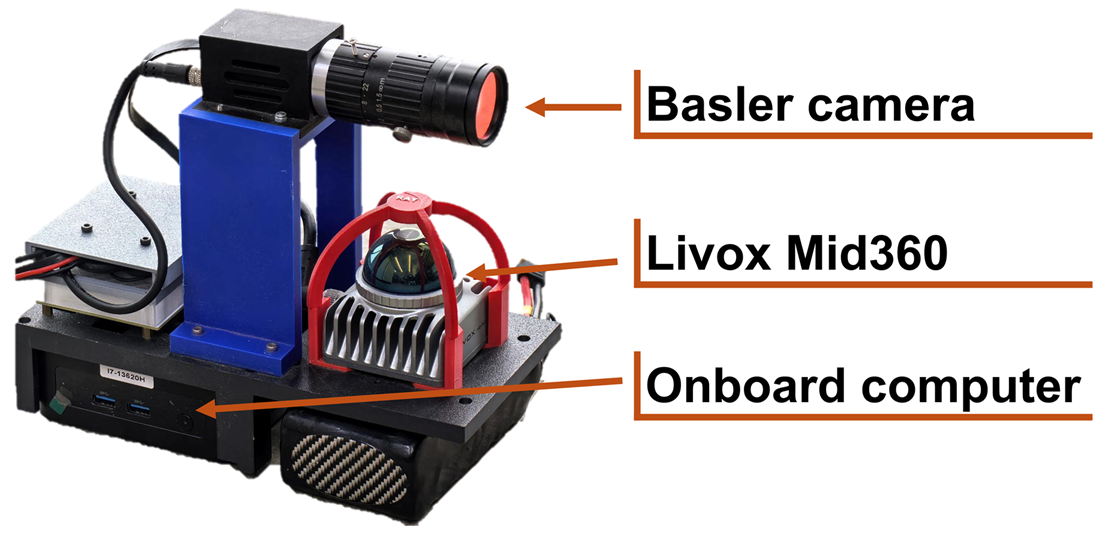}
	\caption{Experimental platform used for data collection}
	\label{fig:experimental_platform}
\end{figure}

The experimental platform consists of a Livox Mid360 LiDAR with its built-in IMU, a monocular Basler camera, and an onboard computer equipped with an Intel i7-13820H CPU and no discrete GPU, as shown in Fig.~\ref{fig:experimental_platform}. All data were collected in real spillway-tunnel environments with strong geometric degeneracy. The evaluation focuses on longitudinal motion along the weakly constrained tunnel direction, where LiDAR scan matching provides limited geometric information. This target condition differs from ordinary indoor corridor degradation because the tunnel cross-section remains nearly self-similar over long distances, as illustrated in Fig.~\ref{fig:tunnel_environment_map}(b) and (c), causing practical loss of LiDAR observability along the tunnel axis. To the best of our knowledge, existing public datasets do not provide such extreme long-range tunnel degeneracy with synchronized LiDAR, camera, and inertial measurements. Since motion-capture or survey-grade full-trajectory ground truth is difficult to deploy in these large-scale industrial tunnels, sparse landmarks placed at known intervals are used as the longitudinal reference for the degenerate direction. This reference is not intended to report full SE(3) accuracy, but directly evaluates whether each method can recover the tunnel-direction motion addressed in this work. A supplementary video documenting the experimental scenes and
landmark passages is available at \url{https://youtu.be/wk4OWfhpPH8}.

We compare the proposed method with FAST-LIVO2~\cite{zheng2024fastLIVO2}, R3LIVE~\cite{lin2022r3live}, FAST-LIO2~\cite{xu2022fastLIO2}, DLIO~\cite{chen2023DLIO}, and VINS-Mono~\cite{qin2018vins}. FAST-LIO2 and DLIO represent LiDAR-inertial odometry without visual compensation, while VINS-Mono provides a monocular visual-inertial baseline. FAST-LIVO2 and R3LIVE are representative tightly coupled LiDAR-inertial-visual baselines. Their inclusion examines whether general-purpose visual integration is sufficient to recover tunnel-direction motion without explicitly conditioning visual measurements and state updates on the LiDAR-deficient subspace. Two real sequences are used. In the first sequence, the sensor unit in Fig.~\ref{fig:experimental_platform} was mounted on the cart platform in Fig.~\ref{fig:cart_spillway}(b) and moved approximately along the tunnel direction while facing the wall. In the second sequence, the same unit was handheld and moved along the tunnel with an oblique view toward the wall, introducing stronger motion disturbance and a different viewing geometry.

\subsection{Cart-Mounted Spillway-Tunnel Evaluation}
\label{subsec:cart_spillway}

\begin{figure*}[t]
	\centering
	\includegraphics[width=0.95\textwidth]{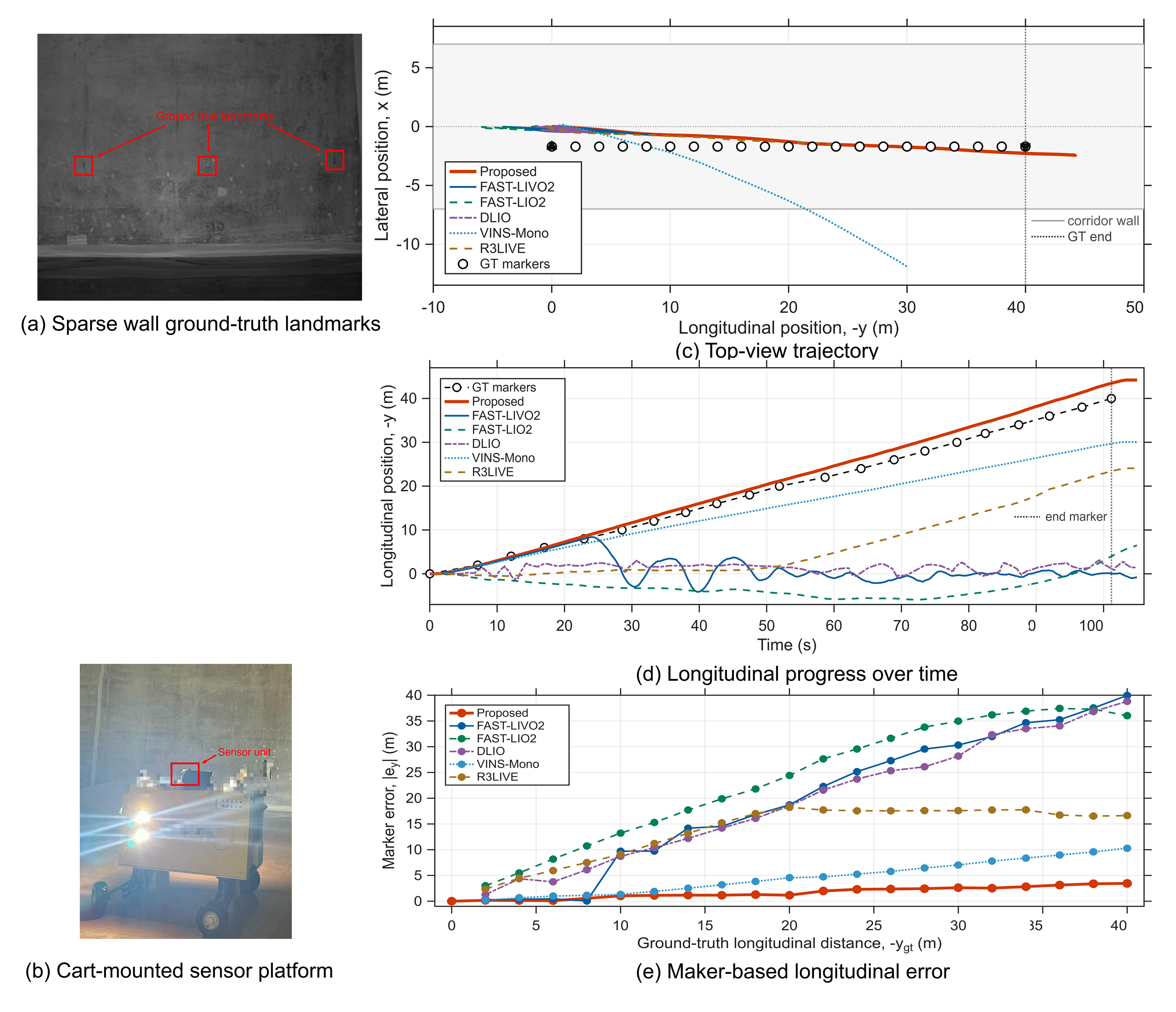}
	\caption{Cart-mounted spillway-tunnel evaluation. (a) Sparse wall landmarks used for longitudinal reference. (b) Sensor platform and experimental scene. (c) Top-view trajectory. (d) Longitudinal progress over time. (e) Marker-based longitudinal error}
	\label{fig:cart_spillway}
\end{figure*}

The cart-mounted sequence provides the main quantitative evaluation. As shown in Fig.~\ref{fig:cart_spillway}(a), sparse wall landmarks were placed at 2~m intervals along the tunnel direction. The passing time of each landmark was manually recorded and used as a sparse longitudinal reference. The landmarks are small in the camera image and only provide reference timestamps for evaluation.

The top-view trajectories in Fig.~\ref{fig:cart_spillway}(c) illustrate the different failure modes. The proposed method follows the expected forward direction while maintaining limited lateral drift. FAST-LIO2~\cite{xu2022fastLIO2} and DLIO~\cite{chen2023DLIO} provide little longitudinal displacement because the LiDAR geometry is weakly constrained along the tunnel axis. FAST-LIVO2~\cite{zheng2024fastLIVO2} initially tracks the motion but subsequently oscillates near the starting region. R3LIVE~\cite{lin2022r3live} maintains a comparatively bounded lateral trajectory but covers only part of the tunnel, indicating that its general tightly coupled formulation does not recover sufficient longitudinal motion. VINS-Mono~\cite{qin2018vins} produces a continuous trajectory, but its pronounced lateral curvature indicates an unreliable attitude estimate in this weak-texture and repetitive scene.

The longitudinal progress in Fig.~\ref{fig:cart_spillway}(d) provides a direct comparison along the degenerate direction. The proposed method produces a monotonic estimate and reaches 43.46~m when the final marker reference is 40~m. R3LIVE~\cite{lin2022r3live} recovers part of the forward motion but reaches only 23.39~m, corresponding to substantial longitudinal under-travel. FAST-LIO2~\cite{xu2022fastLIO2} and DLIO~\cite{chen2023DLIO} remain near the initial longitudinal position or drift backward, while FAST-LIVO2~\cite{zheng2024fastLIVO2} exhibits persistent oscillation and fails to recover continuous forward motion. VINS-Mono~\cite{qin2018vins} estimates approximately 30~m of progress, but its curved top-view trajectory shows that this longitudinal estimate is accompanied by considerable full-state drift.

For quantitative evaluation, the marker-based longitudinal error is computed at the marker timestamps as
\begin{equation}
	|e_y| = |y_{\mathrm{est}} - y_{\mathrm{gt}}| .
	\label{eq:marker_longitudinal_error}
\end{equation}
As shown in Fig.~\ref{fig:cart_spillway}(e), the proposed method maintains the smallest error with respect to the sparse marker references. The errors of FAST-LIVO2~\cite{zheng2024fastLIVO2}, FAST-LIO2~\cite{xu2022fastLIO2}, and DLIO~\cite{chen2023DLIO} increase with traveled distance, indicating that their longitudinal motion estimates are not recovered under this degeneracy. R3LIVE~\cite{lin2022r3live} recovers more forward motion than these failed baselines, but its persistent under-travel produces a final longitudinal error of 16.61~m. Although VINS-Mono~\cite{qin2018vins} has a smaller longitudinal error than R3LIVE, its curved top-view trajectory indicates that its full-state estimate remains unreliable.

\begin{table}[t]
	\centering
	\caption{Marker-Based Longitudinal Accuracy in the Cart-Mounted Spillway-Tunnel Sequence}
	\label{tab:cart_marker_accuracy}
	\footnotesize
	\setlength{\tabcolsep}{4pt}
	\begin{tabular}{lcccc}
		\hline
		Method & RMSE & $|e_{\mathrm{end}}|$ & Final progress & Status \\
		& (m) & (m) & ratio (\%) & \\
		\hline
		\textbf{Proposed} & \textbf{2.00} & \textbf{3.46} & \textbf{91.4} & complete \\
		FAST-LIVO2~\cite{zheng2024fastLIVO2} & 23.84 & 39.96 & 0.1 & oscillatory \\
		FAST-LIO2~\cite{xu2022fastLIO2} & 26.62 & 36.01 & 10.0 & degenerate \\
		DLIO~\cite{chen2023DLIO} & 22.96 & 38.82 & 3.0 & degenerate \\
		VINS-Mono~\cite{qin2018vins} & 5.67 & 10.29 & 74.3 & curved drift \\
		R3LIVE~\cite{lin2022r3live} & 14.76 & 16.61 & 58.5 & under-travel \\
		\hline
	\end{tabular}
\end{table}

Table~\ref{tab:cart_marker_accuracy} summarizes the marker-based accuracy. The final progress ratio is defined as
\begin{equation}
	r_{\mathrm{prog}}
	=
	\max\left(0, 1-\frac{|e_{\mathrm{end}}|}{40~\mathrm{m}}\right)
	\times 100\% .
	\label{eq:final_progress_ratio}
\end{equation}
The proposed method obtains the lowest RMSE of 2.00~m and the lowest final error of 3.46~m, with a final progress ratio of 91.4\%. R3LIVE~\cite{lin2022r3live} reaches 58.5\% of the reference progress but retains a final error of 16.61~m, while VINS-Mono~\cite{qin2018vins} reaches 74.3\% longitudinal progress but suffers from curved drift. FAST-LIVO2~\cite{zheng2024fastLIVO2}, FAST-LIO2~\cite{xu2022fastLIO2}, and DLIO~\cite{chen2023DLIO} remain at or below 10.0\%. Overall, the cart-mounted results show that the proposed method provides the most accurate and stable longitudinal estimate in this long-range, highly degenerate sequence, whereas the evaluated LIO, LIVO, and VIO baselines exhibit stagnation, oscillation, under-travel, or full-state drift.

\subsection{Handheld Sequence, Controlled Ablations, and Degeneracy-Spectrum Analysis}
\label{subsec:handheld_spillway}

\begin{figure}[t]
	\centering
	\includegraphics[width=\columnwidth]{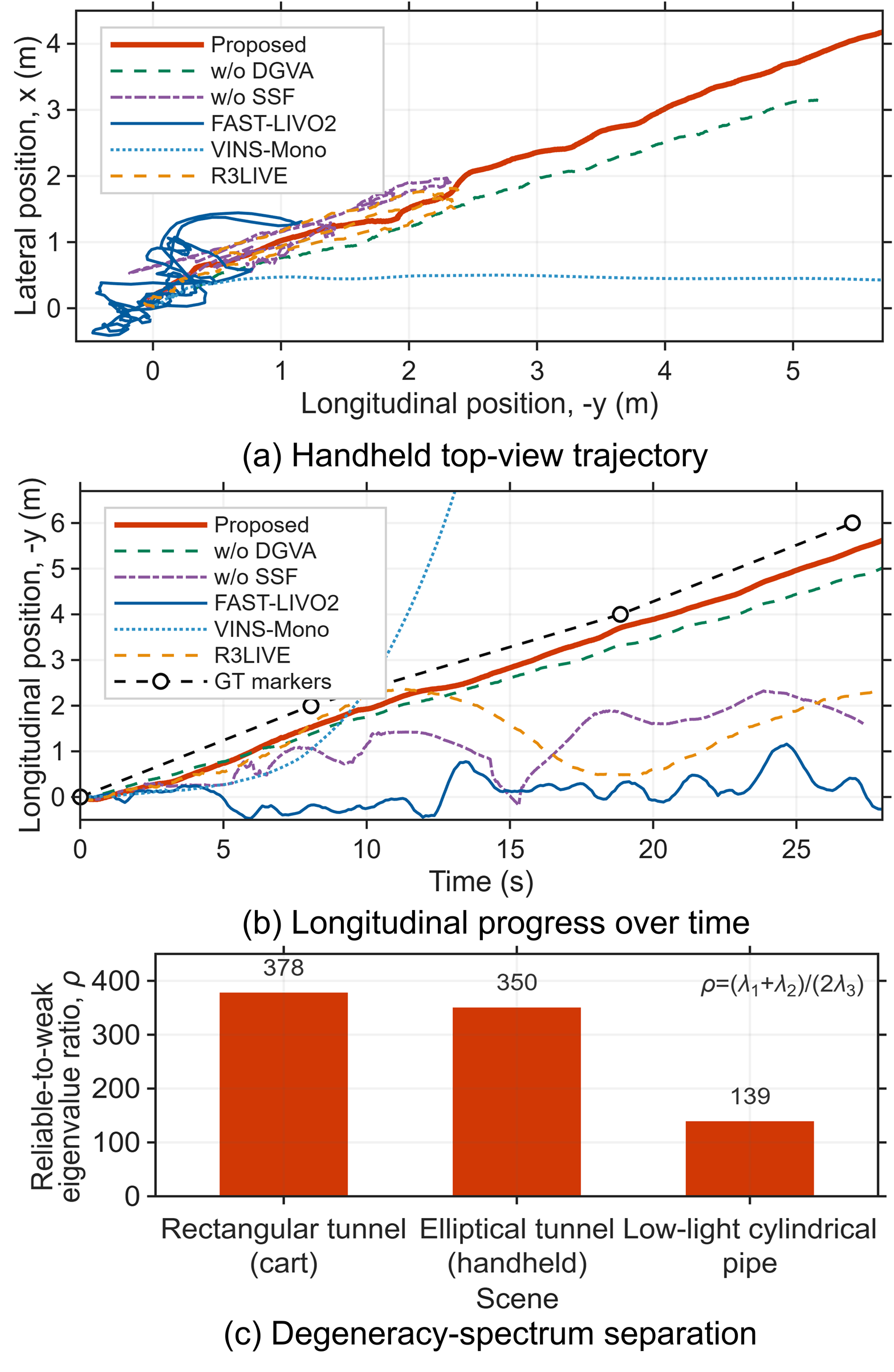}
	\caption{Handheld oblique sequence and degeneracy-spectrum analysis. (a) Handheld top-view trajectory. (b) Longitudinal progress over time. (c) Degeneracy-spectrum separation. The cylindrical pipe sequence is used only for spectrum analysis because its image contrast is too low for reliable visual compensation}
	\label{fig:handheld_spillway}
\end{figure}

The handheld sequence evaluates the method under less constrained acquisition. The sensor unit was carried by hand, producing stronger motion disturbance than the cart-mounted sequence, and the camera viewed the wall obliquely rather than frontally. The marker reference is used only as an approximate longitudinal reference because oblique viewing and manual recording introduce larger uncertainty. Since FAST-LIO2~\cite{xu2022fastLIO2} and DLIO~\cite{chen2023DLIO} exhibit the same LiDAR-degenerate behavior as in the cart-mounted sequence, Fig.~\ref{fig:handheld_spillway} focuses on FAST-LIVO2~\cite{zheng2024fastLIVO2}, R3LIVE~\cite{lin2022r3live}, VINS-Mono~\cite{qin2018vins}, and the two controlled ablation variants.

The first ablation, w/o Degeneracy-Guided Visual Association (w/o DGVA), replaces the proposed degeneracy-guided patch selection and LiDAR-anchored persistent tracking with conventional feature detection and optical-flow tracking, while keeping the remaining fusion pipeline unchanged. The second ablation, w/o Subspace-Selective Fusion (w/o SSF), keeps the proposed visual residual construction but removes the subspace-selective fusion. The visual residuals are then applied as ordinary full-state pose constraints. These two variants isolate the effects of the proposed visual association strategy and the proposed subspace fusion strategy, respectively.

Fig.~\ref{fig:handheld_spillway}(a) and Fig.~\ref{fig:handheld_spillway}(b) show the top-view trajectory and longitudinal progress. The proposed method remains stable and follows the expected tunnel direction with a monotonic longitudinal estimate. FAST-LIVO2~\cite{zheng2024fastLIVO2} stays near the starting region with local oscillation, consistent with its cart-mounted failure mode. R3LIVE~\cite{lin2022r3live} initially follows the forward motion but subsequently loses much of the accumulated longitudinal progress and finishes at only about 2.3~m for the approximately 6~m reference. VINS-Mono~\cite{qin2018vins} diverges after the stronger handheld motion begins. The w/o DGVA variant underestimates the forward progress, indicating that generic feature and optical-flow associations introduce unreliable visual residuals in weak-texture and repetitive tunnel images. The w/o SSF variant exhibits a substantially larger position bias because its visual residuals are forced to constrain the full six-degree-of-freedom pose. Together with the controlled ablations, the handheld results show that the proposed method remains effective under stronger motion disturbance and oblique viewing, and that both degeneracy-guided visual association and subspace-selective fusion contribute to this robustness.

To further examine the role of LiDAR subspace masking, Fig.~\ref{fig:handheld_spillway}(c) compares the spectral separation between reliable and weak LiDAR directions using
\begin{equation}
	\rho =
	\frac{\lambda_1+\lambda_2}{2\lambda_3},
	\label{eq:degeneracy_spectrum_ratio}
\end{equation}
where $\lambda_1$ and $\lambda_2$ denote the information magnitudes of two relatively reliable directions and $\lambda_3$ denotes the weakest direction. A larger $\rho$ indicates a clearer separation between the well-constrained and degenerate LiDAR subspaces. In the two trajectory-evaluation sequences, this separation is already strong, so the unreliable LiDAR components along the weak direction have limited influence on the final trajectory after visual compensation. Consequently, the trajectory-level effect of LiDAR subspace masking is less visually pronounced in these cases.

However, this behavior is not universal across tunnel geometries. Different cross-sectional shapes and point-cloud distributions can produce substantially different degeneracy spectra. The low-light cylindrical pipe sequence shows a weaker separation between reliable and degenerate directions. In such cases, residual LiDAR Hessian and gradient components along the weak direction are more likely to introduce noisy back-and-forth corrections and interfere with visual compensation. Although this pipe sequence is not used as a full trajectory benchmark due to insufficient image contrast, it illustrates why LiDAR subspace masking remains necessary as a safeguard against unreliable LiDAR information in the deficient direction.

Taken together, the two trajectory evaluations provide complementary evidence for the proposed framework. The cart-mounted sequence provides a long-range quantitative assessment under comparatively constrained platform motion, whereas the handheld sequence examines robustness to stronger motion disturbance and oblique viewing and further isolates the contributions of the visual association and subspace fusion components. Across these distinct acquisition conditions, the proposed method consistently recovers stable tunnel-direction progress, while the evaluated generic LIO, LIVO, and VIO approaches exhibit different forms of longitudinal failure or full-state drift. These complementary results support the effectiveness of conditioning both visual constraint construction and state estimation on the LiDAR-deficient subspace in the evaluated tunnel environments.

\section{Conclusion}
\label{sec:conclusion}

 This paper presented DGV-DLIVO, a degeneracy-guided LiDAR-visual-inertial odometry framework for continuous localization in long hydropower passages with severe geometric and visual degradation. By exploiting the LiDAR weak direction as an active guidance signal, DGV-DLIVO jointly organizes visual information selection, LiDAR-visual metric association, and subspace-selective fusion. Unlike conventional multimodal odometry methods that introduce vision as a general complementary modality, the proposed framework enables targeted recovery of the motion component lost by LiDAR while preserving reliable geometric constraints.

Real-world experiments in severely degenerate spillway tunnels demonstrate the robustness advantage of DGV-DLIVO over representative state-of-the-art LIO, LIVO, and VIO methods. Under the same challenging conditions, existing approaches suffer from stagnation, oscillation, severe under-travel, curved drift, or divergence, whereas DGV-DLIVO maintains continuous and directionally consistent tunnel-axis motion estimation under both cart-mounted and handheld acquisition. It achieves a 91.4\% progress ratio in the 40~m cart-mounted sequence, substantially outperforming all evaluated baselines in tunnel-axis motion recovery and trajectory stability. Ablation studies further verify the effectiveness of degeneracy-guided visual constraint construction and subspace-selective adaptive fusion.

DGV-DLIVO provides a direction-aware multimodal estimation strategy for robotic systems operating in structurally degenerate environments, offering a reliable localization foundation for long-range autonomous inspection in repetitive underground infrastructures.

\backmatter

\bmhead{Acknowledgements}

The authors thank the facility operator for providing access to the experimental
site and facilitating data collection.

\section*{Statements and Declarations}

\bmhead{Funding}

No funding was received for conducting this study.

\bmhead{Competing Interests}

The authors have no relevant financial or non-financial interests to disclose.

\bmhead{Data availability}

The datasets generated and analyzed during the current study are not publicly
available due to confidentiality agreements and access restrictions associated
with the operating facility.

\bmhead{Code availability}

Source code and a supplementary video are available from the anonymous project
page at \url{https://github.com/0RayZhang0/DGV-DLIVO}.

\bmhead{Author contributions}

The author contribution statement has been omitted from the manuscript to preserve anonymity during double-blind peer review and will be provided after the review process.

\bibliography{references}

\end{document}